\documentclass[conference]{IEEEtran}

\usepackage[T1]{fontenc}
\usepackage{lmodern}
\usepackage{amssymb}
\usepackage{amsfonts}
\usepackage{amsmath}
\usepackage{graphicx}
\usepackage{booktabs}
\usepackage{xcolor}
\usepackage{url}
\usepackage{siunitx}
\usepackage{hyperref}

\newcommand{\ZZ}{\mathbb{Z}}
\newcommand{\QQ}{\mathbb{Q}}
\newcommand{\RR}{\mathbb{R}}
\newcommand{\OK}{\mathcal{O}_K}

\newcommand{\R}{\mathcal{R}}
\newcommand{\bigO}[1]{O(#1)}
\newcommand\mat[1]{#1}

\newcommand{\QINT}{\mathsf{QINT}}

\newcommand{\ceil}[1] {\ensuremath{{\lceil  #1 \rceil }}}
\newcommand{\rx}[1]{\ensuremath{#1\!\times}}

\begin{document}

\title{Algebraic Integer Quantization for Machine Learning Applications}


\author{\IEEEauthorblockN{Kostal-Vazquez}
\IEEEauthorblockA{\textit{Lemurian Labs Research Group} \\
Toronto, Canada \\
Anthony@lemurianlabs.com}

\and
\IEEEauthorblockN{Dimitrov}
\IEEEauthorblockA{\textit{Lemurian Labs Research Group} \\
\textit{University of Calgary}\\
Calgary, Canada \\
Vassil@lemurianlabs.com}



\and
\IEEEauthorblockN{Butler}
\IEEEauthorblockA{\textit{Lemurian Labs Research Group} \\
United States of America \\
Branden@lemurianlabs.com}
}
\maketitle

\begin{abstract}
  In this article, we propose algebraic integer number formats (AINF)
  as a framework for highly configurable and accurate compressed
  computation. A key property of AINF is its straightforward reduction to
  purely integer arithmetic, enabling error-free multiply-accumulate operations.
  We develop a matrix multiplication engine that exploits this property, obtaining
  performance within a predictable factor of the most efficient numerical counterparts
  on modern hardware accelerators. Another key advantage of AIQ is the freedom to select
  algebraic integers tailored to the numerical characteristics of a given use case, offering
  optimization opportunities that are fundamentally unavailable in floating-point
  or fixed-point formats. Motivated by these properties, we introduce
  algebraic integer quantization (AIQ), a lossy compression format for large language model (LLM)
  inference, and provide results demonstrating that AIQ achieves competitive accuracy
  compared to the industry-standard MX format across a diverse set of models and
  benchmarks at equivalent bit-per-parameter budgets. The experiments presented in this
  article employ algebraic integers of degree two (quadratic integers); generalizations
  to higher-order and, in particular, multidimensional algebraic integers promise
  even greater computational advantages and remain a promising direction for future work.
\end{abstract}


\section{Introduction}

The set of algebraic integers, defined as roots of monic polynomials
of degree two or higher with integer coefficients, plays a vital role
in number theory. From a purely algorithmic point of view, the fact
that the basic arithmetic operations (additions, subtractions and
multiplications) can be done in an error-free manner makes the
algebraic integers a promising data format for applications requiring
highly reliable numerical results.  Algebraic integers have a rich
history in engineering applications. In this article, we explore their potential for machine learning.

To the best of our knowledge, Irving Reed was the first one who
proposed the use of algebraic integers for error-free computation of
convolutions in 1976~\cite{1055583}. Next in the early 80s, Dubois and
Venetsanopoulus showcased how the use of algebraic integer
quantization can be used for implementing FFTs of various lengths (not
necessarily powers of two)~\cite{1163370}. In 1985, Cozzens and
Finkelstein proposed the idea to combine algebraic integer
quantization with residue number systems for highly parallel
implementation of FFTs~\cite{1057081}. Their idea constitutes the use
of two levels of parallelism: level one, based on the use of algebraic
integers and, level two, based on the use of residue number systems
which can be viewed as a generalization of the quadratic residue
number system. In the next few years researchers from MITRE
corporation performed many investigations on that. Some papers from
that era and some recently declassified reports provide a good idea
about the depth of these efforts~\cite{MITRE_AIQ,101978}.

Since 1998~\cite{681703}, it has been discovered that algebraic
integers can offer significant advantages for implementing the
discrete cosine transform (DCT), which is the most important (and time
consuming) part of several image compression standards. The
introduction in 2003~\cite{doi:10.1049/el:20030406} of the concept of
multidimensional algebraic integer quantization led to many additional
improvements. Two ARITH articles~\cite{1207684,1467642} showcase the
specifics of these encoding schemes and the associated computational
benefits.

Algebraic Integer Number Format (AINF) encoding differs significantly
from traditional floating-point formats. For example, it provides a
single exact representation of zero, can accommodate both positive and
negative numbers without requiring a dedicated sign bit and its
arithmetic satisfies associativity and distributivity.  In AINF, real
numbers are represented by a tuple of integers. The arithmetic
operation only involves integer arithmetic and is therefore error
free. AINF is a mixed precision format, i.e. the total bit length of
the output of an addition or a multiplication may be larger than the
bitlength of the inputs.  This should not be seen as a major obstacle
since modern hardware provide many instructions for efficient mixed
precision arithmetic, both with integer or floating-point operands.

In this article, we report results on the use of algebraic integer encodings for ML applications along two independent lines of investigation. First, we describe algorithms for matrix multiplication over algebraic integers and their optimized implementation using dedicated instructions available on modern hardware accelerators, demonstrating that exact integer-only matrix multiplication is feasible at competitive performance levels. Second, we provide end-to-end large language model (LLM) quantization experiments comparing AINF against standard quantization formats at the 8-bit and 6-bit levels, evaluating the representational quality of the algebraic integer format independently of its arithmetic properties. Combining these capabilities into a unified compute and compression scheme is a natural next step. Our aim is to renew interest in algebraic integer arithmetic within the computer arithmetic community. As low-bitlength data formats (FP8, FP4) become the norm in modern LLM architectures, we believe the flexibility of AIQ parameter selection provides optimization opportunities that simply do not exist in floating-point or fixed-point encodings.

\section{Algebraic and Quadratic Integers}
\label{sec:algebraic-integers}

An \emph{algebraic integer} is a complex root of some monic
polynomial\footnote{A univariate polynomial whose leading coefficient
  is $1$.} with integer coefficients. Given an algebraic integer
$\alpha$, there exists a unique monic polynomial $f \in \ZZ[X]$ of
least degree such that $f(\alpha)=0$. This polynomial is called the
\emph{minimal polynomial} of $\alpha$. Moreover, if this minimal
polynomial has degree $n$, then $\alpha$ is said to be of degree $n$.

A \emph{quadratic integer} is an algebraic integer of degree 2. For
example $\sqrt{2}$ is a quadratic integer because it is a root of
$x^2-2$.

The set of \emph{all} quadratic integers is not a ring because it is
not closed under addition and multiplication\footnote{as e.g.
  $\sqrt{2} + \sqrt{3}$ is not a root of a degree-2 polynomial.}.
Thus, in order to be able to do additions and multiplications, we
consider the set of quadratic integers that belong to a given
\emph{quadratic number field} $K = \QQ(\sqrt{D})$, where $D$ is a
square-free integer called the \emph{discriminant} of $K$.
For any such quadratic number field, this set is
called the \emph{ring of integers} of $K$, and is denoted $\OK$. The
elements of $\OK$ may be written $a + b\cdot\omega$, where
$a,b \in \ZZ$, and $\omega$ is an algebraic integer defined by:
\begin{equation}
  \label{eq:omega}
  \omega = 
  \begin{cases}
    \sqrt{D} &\text{if } D \equiv 2,3 \pmod{4}\\
    (1+\sqrt{D})/2 &\text{if } D \equiv 1 \pmod{4}\\
  \end{cases}
\end{equation}

In this paper, we are studying the use of quadratic integers
$a + b\cdot\omega$ to represent finite subsets of $\RR$.
Therefore, we only consider \emph{real} quadratic number fields (i.e.
$D > 0$) and we fix the bitwidth of the integer coefficients $a,b$.
In order to represent both positive and negative numbers, we consider
that $a$ and $b$ are $m$-bit integers encoded using two's complement.
Since the elements of $\OK$ are entirely determined by $D$, we shall
use the notation
\begin{multline}
\QINT_{2m}(D) = \\
\left\{a + b\cdot\omega \;;\; a,b \in \{-2^{m-1},\dots, 2^{m-1}-1\}\right\},
\end{multline}
where $\omega$ is given by~(\ref{eq:omega}).

In general, note that $a$ and $b$ do not need to have the same
bitlength. In this case, we shall use the notation $\QINT_{m,n}(D)$.

\subsection{Arithmetic operations}
\label{sec:arith-ops}

Let $(a,b)$ and $(a',b')$ two quadratic integers from
$\QINT_{2m}(D)$.
For the addition, we have:
$$
(a,b) + (a',b') = (a+a', b+b').
$$
For the multiplication, let $\pi(x) = x^2 - \pi_1 x - \pi_0$ be the
minimal polynomial of $\omega$. Then, since $\pi(\omega) = 0$, we
have $\omega^2 = \pi_1 \omega + \pi_0$, and thus
$$
(a,b) \times (a',b') = (aa' + \pi_0bb', ab' + a'b + \pi_1bb').
$$
It is important to notice that the $\QINT$ format requires mixed
precision arithmetic since the result of an addition or a
multiplication might belong to $\QINT_{2\overline{m}}(D)$
with $\overline{m} > m$. As we shall see later, this can be handled
nicely with the tensor core instructions that also use mixed precision
arithmetic.
  
For example, the 8-bit type
$$\QINT_{8}(2) = \left\{a + b\cdot\sqrt{2} \;;\; a, b \in
  \{-8,\dots,7\} \right\}$$ can be used to represent 256 real numbers
between $-19.3137$ and $16.8994$. The pair $(0,0)$ is the unique exact
representation of zero. The smallest non-zero representatives are
$\pm 0.071$. The addition of two elements from $\QINT_8(2)$ is of type
$\QINT_{10}(2)$, while their product is of type $\QINT_{10,9}(2)$ since
we have $\pi_1=0, \pi_0=2$.


\section{Matrix multiplication  over algebraic integer}
\label{sec:aiq-matmul}
Matrix multiplication is the cornerstone operation in the pipeline of
both the inference and learning phases arising in machine learning
algorithms. Leveraging the use of optimized hardware accelerators
units for that operation, such as \emph{Tensor core} on Nvidia or
\emph{Matrix core} on AMD, is nowadays mandatory to reach exascale
computation required by AI applications. The current performances of
those (fixed small-size) \emph{Matrix Multiplication Units (MMU)} on a
single modern GPU are already in the petaflopic range, as illustrated
in Table~\ref{table:MMUPerf}.
 
\begin{table}[!h]
  \caption{Performance of matrix multiplication units on modern
    GPUs, in TFLOPS (or TOPS for INT8).}
    \label{table:MMUPerf}
    \centering
    \begin{tabular}{lcccc}
      \toprule
      Architecture & FP32 & BF16 & FP8 & INT8 \\
      \midrule
      AMD MI300X  & 490.3 & 1307.4 & 2614.9 & 2614.9 \\
      NVIDIA Blackwell B200 & 2200 & 4500 & 9000 & 9000 \\
      \bottomrule
    \end{tabular}
\end{table}

These units natively support mixed-precision arithmetic, accumulating products in a larger type such as FP32 for floating-point formats, or INT32 for integer formats. This makes them well suited for AINF computation.

As already mentioned in Section~\ref{sec:algebraic-integers}, our
$\QINT_{2m}(D)$ format requires mixed precision arithmetic. We will
show that it can be done easily through these mixed precision matrix
multiplication units.

We do not claim that the following result are new, since such a
solution is well known from the relation of algebraic integer ring and
polynomial quotient ring. In particular, many similarities arise in the
computation of complex matrix multiplication since it corresponds to
the imaginary number field $\QQ(\sqrt{-1})$
(see~\cite{uchino2025emulationcomplexmatrixmultiplication}).

Let $K=\QQ(\sqrt{D})$ with $D>0$ a real quadratic number field and
$\OK$ its ring of integers.  For simplicity, we shall use $\R$ instead
of $\OK$ in the following.
We have
$\R = \{\alpha + \omega \cdot \beta \;;\; \alpha,\beta \in \ZZ\}$,
where $\omega$ is a root of the polynomial
$\pi(x)=x^2-\pi_1x-\pi_0 \in \ZZ[x]$.

Given two matrices $\mat{A},\mat{B} \in \R^{n \times n}$, our
algebraic integer matrix multiplication engine aims at computing
$\mat{C}=\mat{A}\times \mat{B} \in \R^{n \times n}$ using the minimal
number of integer operations, while allowing the use of
\emph{MMU}. Since $\R$ is isomorphic to $\ZZ[x]/\pi(x)$, the matrices
$A$ and $B$ can be represented by the polynomial matrices
$\bar{A}(x)=A_0+x A_1$ and $\bar{B}(x)=B_0 + x B_1$ in
$\ZZ^{n \times n}[x]$, such that $A=\bar{A}(\omega)$ and
$B=\bar{B}(\omega)$. Therefore, computing the product $C=A\times B$
reduces to a few polynomial matrix multiplications, followed by a
reduction modulo $\pi(x)$, yielding the
equation:
\begin{equation}\label{eq:aiq-matmul1}
  C = (A_0B_0 + \pi_0 A_1B_1) + \omega \cdot (A_0B_1+A_1B_0 + \pi_1 A_1B_1)
\end{equation}

Equation~(\ref{eq:aiq-matmul1}) can be reformulated as a single matrix
multiplication of dimension $(2n \times 2n) \times (2n \times n)$, as
follow:
\begin{equation}\label{eq:aiq-matmul2}
  \begin{pmatrix}
    C_0\\
    C_1\\
  \end{pmatrix}
  =
  \begin{pmatrix}
     A_0 & \pi_0 A_1 \\
     A_1 & A_0+\pi_1A_1\\
   \end{pmatrix}
   \times
  \begin{pmatrix}
    B_0\\
    B_1\\ 
  \end{pmatrix},
\end{equation}
where $C=C_0+\omega C_1$. It is clear from these equations that the
asymptotic complexity of both approaches is dominated by the matrix
product since the pre (or post) matrix linear combinations involving
$\pi_0$ and $\pi_1$ only incur a quadratic cost. To further decrease
algorithmic complexity, i.e the number of $n\times n$ matrix multiplications,
one could employ any sub-quadratic polynomial multiplication
algorithm. In particular, Karatsuba's approach~\cite{KarOfm63} allows
to reduce the number of matrix multiplication in
Equation~\ref{eq:aiq-matmul1} to only three, using
\begin{equation}
  C= (\mat{\Delta_{00}}  + \pi_0 \cdot \mat{\Delta_{11})} +\omega \cdot (\mat{\Delta_{01}} - \mat{\Delta_{00}}- \mat{\Delta_{11}}+ \pi_1\cdot \mat{\Delta_{11}})  \label{eq:matmul_kara_qint} 
\end{equation}
where $\mat{\Delta_{00}}= \mat{A_0}\mat{B_0}$,
$\mat{\Delta_{11}}= \mat{A_1}\mat{B_1}$,
$\mat{\Delta_{01}}= (\mat{A_0}+\mat{A_1}) \times
(\mat{B_0}+\mat{B_1}).$ One may again notice that all the linear
combinations of the matrices will remain in $\bigO{n^2}$, yielding a
theoretical improvement of $25\%$, i.e. one matrix multiplication is
saved.

As already mentioned in Section~\ref{sec:arith-ops}, mixed precision
is required to handle the arithmetic in $\R$. To ease
the presentation, we now assume that $\R_t=\QINT_{t}(D)$, and
$\lambda_0= \ceil{\log_2(\pi_0)}$, $ \lambda_1=
\ceil{\log_2(\pi_1)}$. Given $z,z' \in \R_{2m}$, we then have:
\begin{align}
  z + z' \in \R_{2t},& \text{ with } t \leq  m+1 \\
  z\times z' \in \R_{2t},& \text{ with } t \leq 2m+ \max(1+\lambda_0,2+\lambda_1)
\end{align}

The above inequalities allow to define the right algebraic integer
format of a scalar product. Let $\vec{u}, \vec{v} \in (\R_{2m})^n$, we
then have:
$$
  \langle \vec{u}, \vec{v} \rangle \in \R_{2t} \text{ with } t \leq 2m+ \max(1+\lambda_0,2+\lambda_1) + \log n
$$

Targeting the use of INT8 \emph{Matrix Multiplication Units} that accumulate 8-bit integer products over 32-bit
integers, we need $n$ to satisfy:
$$
  \log n \leq 32 - 2m - \max(1+\lambda_0,2+\lambda_1)
$$

For instance, taking entries in $\QINT_{16}(2)$, we are able to handle
matrix multiplication of inner dimension at most $n = 2^{14} =
16384$. For $\QINT_{8}(2)$, this increases to inner dimension
$n = 2^{22}\approx 4\cdot 10^6$. Changing the discriminant $D=5$,
leading $\omega$ to be the golden ratio, i.e. $\omega=(1+\sqrt{5})/2$,
reaches the exact same dimensions as $\pi_0=\pi_1=1$. Nevertheless,
the latter choice of $\QINT_{2m}(5)$ allows some saving within
Karatsuba's approach. Indeed, in that case,
Equation~\ref{eq:matmul_kara_qint} becomes
$ C= (\mat{\Delta_{00}} + \pi_0 \cdot \mat{\Delta_{11})} +\omega \cdot
(\mat{\Delta_{01}} - \mat{\Delta_{00}})$.

We shall mention that the usage of Karatsuba includes some limitation
due to the pre-linear combination that can carry an extra bit. If
$\QINT_{2m}$ is such that $m<8$, this extra bit does not incur any
overflow, and using \emph{MMU} $8\times8 \mapsto 32$ is safe, but it is
not true for $m=8$.

\subsection{Experimental analysis}

\begin{table*}[!ht]
\centering
\caption{Performance of matrix multiplication on AMD MI300X for square matrices (in TOPS), with slowdown factor for
  $\QINT$ }
\label{tab:bench_matmul}
\bigskip
\sisetup{round-mode=places, round-precision=1}
\resizebox{0.85\textwidth}{!}{%
\begin{tabular}{r
    S[table-format=4.1]
    S[table-format=4.1]
    S[table-format=4.1]
    S[table-format=3.1] S[table-format=1.2] S[table-format=1.2]
    S[table-format=3.1] S[table-format=1.2] S[table-format=1.2]}
\toprule
  & {INT8} & {BF16} & {FP8} & \multicolumn{3}{c}{$\QINT_{16}(5)$-Karatsuba} & \multicolumn{3}{c}{$\QINT_8(5)$-Karatsuba} \\
\cmidrule(lr){5-7}\cmidrule(lr){8-10}
{Size}
  & {TOPS}
  & {TOPS}
  & {TOPS}
  & {TOPS} & {$/\text{INT8}$} & {$/\text{BF16}$}
  & {TOPS} & {$/\text{INT8}$} & {$/\text{FP8}$} \\
\midrule
$1024$  &   55.9 &   83.5 &   61.9           &  11.2 & \rx{4.97} & \rx{7.42} &  11.3 & \rx{4.96} & \rx{5.48} \\
$2048$  &  189.0 &  325.2 &  315.9           &  47.3 & \rx{3.99} & \rx{6.87} &  47.2 & \rx{4.01} & \rx{6.69} \\
$4096$  &  667.8 &  569.1 &  848.4           & 160.8 & \rx{4.15} & \rx{3.54} & 162.0 & \rx{4.12} & \rx{5.24} \\
$8192$  &  802.7 &  621.1 & 1043.5           & 225.0 & \rx{3.57} & \rx{2.76} & 227.7 & \rx{3.53} & \rx{4.58} \\
$16384$ &  828.1 &  645.5 & 35.8\textsuperscript{$\dagger$} & 231.2 & \rx{3.58} & \rx{2.79} & 255.5 & \rx{3.24} & \rx{0.14}\textsuperscript{$\dagger$} \\
\bottomrule
\end{tabular}%
}
\par
\smallskip
$(\dagger)$ \texttt{hipblasLtMatmul function from HIPBLAS\_LT library suffers from dramatic slowdown above $2^{13}$,
  seems to be known}.
\end{table*}

We conducted a matrix multiplication benchmark on a single AMD
Instinct MI300X accelerator to see whether algebraic integer can
leverage the integer matrix core units. Matrix multiplications were
performed with square matrices of dimensions $n \times n$, ranging
from $1024$ to $16{,}384$ which are typical tensor sizes in machine
learning. Each configuration was run for a sufficient number of
iterations, between $10^2$ and $10^5$ depending on the dimensions, to
ensure statistical stability, preceded by a warmup phase of half the
iteration count to stabilize GPU performances. We report average
wall-clock time per iteration, with explicit L2 cache flushing between
iterations. We compare four families of implementations:
\begin{itemize}
    \item INT8:  8-bit integer GEMM (\texttt{rocblas\_gemm\_ex},)
    \item BF16:  bfloat16 GEMM  (\texttt{rocblas\_gemm\_ex}),
    \item FP8:  float8 (E4M3) GEMM (\texttt{hipblasLtMatmul}),
    \item $\QINT_{m}$: our proposed quadratic integer method, in several variants
          (naïve and Karatsuba), with $m=8,16$.
\end{itemize}

Note that despite $\QINT_8$ could in practice rely on INT4 arithmetic,
the MI300X does not provide such a support within Matrix core
units. We therefore revert to store INT4 component into INT8, and
still rely on the INT8 Matrix core unit.  As we shall see, some benefit
can still be achieved on MI300X since memory transfer involves fewer
bit-flipping when integers in the INT4 range are encoded as INT8, thus
reducing the throttling effect.

Table~\ref{tab:bench_matmul} summarizes the performances for each
implementation, where arithmetic throughput is given in TOPS,
according to the input datatype.

A first observation is that $\QINT$ variants are within a factor of at
most five for very small dimensions, to almost three for larger
ones. The penalty of memory transfers, extra additions and kernel
launching explain this degradation for smaller dimensions. Yet, 
we note that $\QINT_{2m}$ matrix multiplication can reach performances
of $m$-bits integer gemm, and thus can be highly beneficial if integer
matrix core is available.

A second observation, which has been already pointed out
in~\cite{uchino2025emulationcomplexmatrixmultiplication}, is that INT8
gemm is not able to achieve the theoretical peak of 2614 TOPS on
MI300X. The situation is similar for FP8 or BF16 gemm
(see~\ref{table:MMUPerf}) but performance degradation seems to be less
important.
This observation explains why we do not exactly observe a 
slowdown factor of $1.5$ between $\QINT_{16}$ and BF16, and of $3$
between $\QINT_8$ and FP8 as theoretically expected.  Yet, our
performances are perfectly aligned with the practical speedups between
INT8 and the other types (BF16, FP8).

Regarding the similar approach developed in
\cite{uchino2025emulationcomplexmatrixmultiplication}, and the better
performances of INT8 gemm on NVIDIA GPUs, we believe that our approach
can be a good competitor for AINF deployment at a software level. This
also suggests that dedicated hardware designed for $\QINT$ type could
really be an interesting path to investigate.



\section{LLM Experiment Background}
\label{sec:llm-exp}
In the world of machine learning, quantization is the process of converting model weights to a different, typically lower-bitwidth data type, reducing memory footprint or enabling more efficient computation at the cost of losing some numerical information. For
example, quantizing a 16-bit model to an 8-bit representation halves
the storage requirement at the expense of potentially reducing the
model accuracy. This trade-off is viable because the internal
representations encoded in a model's trained weights are inherently
approximation tolerant. Lossy compression of these weights can
preserve task-level accuracy even when individual values are
perturbed. The motivation for quantization is straightforward: data
movement dominates the cost of ML inference, so reducing the size of
the data being moved should reduce both memory requirements and,
ideally, latency. In practice, quantization is valuable for two
reasons. First, in memory-constrained environments such as edge
devices, a model may simply not fit in available memory at full
precision, making quantization a necessity rather than an
optimization. Second, smaller data types hold the promise of faster
computation. However, realizing latency gains in practice is
considerably more difficult than achieving memory
savings. Low-bitwidth computation requires dedicated hardware support
for the target data type, efficient kernel implementations that avoid
bottlenecks in packing, unpacking, or format conversion, and a
workload that is compute-bound rather than memory-bound, conditions
that are not always met simultaneously on current accelerators.  To
evaluate a given quantization scheme prior to deployment, \emph{fake
  quantization} is typically employed. In fake quantization, model
weights remain stored in full precision but are rounded to the nearest
representable value in the target format, ensuring that all
computations faithfully reproduce the numerical effects of
low-bitwidth inference. Fake quantization serves as a critical
test. Indeed, if a quantization format cannot preserve model accuracy when the
only source of error is quantization, then there is no reason to
invest in the considerable hardware and software engineering required
to deploy it. Conversely, strong fake quantization results establish
that a format is worth pursuing further.

We now turn to the question of how to measure an LLM's performance. As
of today, LLM performance is evaluated with the use of
benchmarks. Benchmarks are standardized evaluation suites that test a
model's capabilities through tasks analogous to those found in human
assessments, such as reading comprehension, logical reasoning, and
factual recall. Just like human tests, models are graded with an
accuracy score calculated as the number of correct answers divided by
the total number of questions. You can evaluate a quantization scheme
by comparing benchmark accuracy results before and after
quantization. Note that benchmarking has inherent limitations; a model
may perform well on some benchmarks but poorly on others after
compression. Benchmarks may also be suited better to some models over
others. For this reason, any quantization scheme must be validated
across a diverse set of models and benchmarks to establish confidence
in its generality.

It seems fairly far-fetched that a model's weights can be effectively
compressed by half and remain useful by benchmark standards but in
practice this is regularly
seen~\cite{zhou2025precisionneuralnetworkquantization,askarihemmat2022qregregularizationeffectsquantization}. To
understand why, it helps to recall what is actually being
compressed. A neural network's weights are the learned numerical
parameters that define the model's behavior; during inference, input
data is repeatedly multiplied and transformed by these weights as it
passes through successive layers. Several techniques make compressing
them possible. First, not all layers of a model need to be
quantized. Linear layers, which perform simple matrix multiplications
between their weight matrices and incoming activations, are shown to
be fairly resistant to quantization and regularly account for the
majority of a model's total parameter count, so they can be targeted
specifically while leaving more sensitive components such as
normalization layers and embedding tables in full precision. Another
important strategy is to use Block-Scaled Quantization. This is done
to increase the effective dynamic range of low-bitwidth
representations. In block quantization, tensors are partitioned into
contiguous blocks of values. Then each block receives its own scale
factor based on its numerical properties. The way these scale factors
are calculated can vary widely, a simple strategy is to take the
absolute maximum value of a block and adjust it to the quantization
range. This is very similar to the ``Block-wise $k$-bit Quantization''
strategy used by~\cite{dettmers2023qloraefficientfinetuningquantized}
and seems to be a general technique in the field.


\section{AIQ Fake Quantization Experiments}
\label{sec:aiq-fake-quantization}
We now apply algebraic integer number formats to LLM quantization, a
process we term algebraic integer quantization (AIQ). Below is a
strict definition of how we take a model's weights and transform them
using AIQ.

Let $\mathbf{x}$ denote the vector of values within a single
contiguous block, and let $Q$ be the set of representable values in
the quantized type with maximum representable value $r_{\max} =
\max(Q)$.

\noindent\textbf{Scale factor calculation (per block):} The scale factor is computed as
\begin{equation}
s = \frac{r_{\max}}{\max(|\mathbf{x}|)}
\label{eq:scale_factor}
\end{equation}
This will be referred to as the max scale method.

\smallskip
\noindent\textbf{Quantization:} The quantized representation is
obtained by computing
\begin{equation}
\hat{\mathbf{x}} = \mathrm{round}_Q(s \cdot \mathbf{x})
\label{eq:quant}
\end{equation}
where $\mathrm{round}_Q(\cdot)$ maps each scaled value to the nearest
representable value in $Q$.

\smallskip
\noindent\textbf{Dequantization:} The approximate original values are recovered as
\begin{equation}
\tilde{\mathbf{x}} = \frac{\hat{\mathbf{x}}}{s}
\label{eq:dequant}
\end{equation}

In the sequel, we shall use the notation
$$
\langle\text{QuantizeType ScaleType/BlockSize/ScaleMethod}\rangle,
$$
where:
\begin{itemize}
  \item QuantizeType is the set of representatives 
  \item ScaleType is the type used to encode the scale factors
  \item BlockSize is the number of values contained in a block
  \item ScaleMethod is the method of calculating the scale factor
\end{itemize}


To evaluate AIQ as a quantization scheme, we conducted weights-only
fake quantization experiments on two AIQ configurations,
$\langle\QINT_{8}(2)\,\text{BF16/64/max}\rangle$ and
$\langle\QINT_{6}(2)\,\text{BF16/64/max}\rangle$, and on two MX
configurations, $\langle\text{MXFP8 (E4M3) int8/32/mx}\rangle$ and
$\langle\text{MXFP6 (E2M3) int8/32/mx}\rangle$ as described in the OCP
MX specification~\cite{mx_spec}. The inclusion of MX formats provides
a baseline against a widely accepted and industry-standard
quantization scheme. For each experiment, all linear layer weights
were quantized offline and stored in the original layer's data
type. During inference all linear layer operations were performed
against the pre-quantized weights using the model's native
precision. To ensure the results are accurate and hold generality, we
evaluated across five architecturally diverse models,
Llama-3.1-70B-Instruct, Llama-3.1-8B-Instruct, Llama-3.2-1B,
Qwen2.5-7B-Instruct, and Mixtral-8x7B-v0.1, spanning parameter counts
from 1B to 70B and including both dense and mixture-of-experts
architectures across three distinct model families. We then measured
post-quantization performance using four benchmarks,
Winogrande~\cite{DBLP:journals/cacm/SakaguchiBBC21} (commonsense
reasoning), PIQA~\cite{DBLP:conf/aaai/BiskZLGC20} (physical
intuition), BoolQ~\cite{DBLP:conf/naacl/ClarkLCK0T19} (reading
comprehension), and LAMBADA~\cite{DBLP:conf/acl/PapernoKLPBPBBF16}
(language modeling and next-word prediction), chosen to cover a
representative cross-section of language understanding tasks. While
additional benchmarks such as MMLU, HellaSwag, or ARC could further
strengthen the evaluation, the consistency of our results across 5
models $\times$ 4 benchmarks $\times$ 4 quantization schemes (80
individual measurements) provides reasonable confidence that the
observed trends are robust and not artifacts of a particular
model-benchmark pairing.

It is very difficult to compare different quantization types. However
we can ensure a fair comparison by normalizing the results on a bit
per parameter basis. Bits per parameter is a measure of how many bits
of information are needed in a truly quantized setting. For example in
the $\langle\QINT_{8}(2)\,\text{BF16/64/max}\rangle$ scheme the model
weights would be stored as a pair of 4-bit integers along with a BF16
scale factor for each block of 64 values. This makes the theoretical
bit-per-parameter of this storage format $(16/64 + 8 = 8.25$ bits per
parameter). In Table~\ref{tab:bit_per_param}, we give the number of the
bits per parameter for each scheme we have tested.

\begin{table}
  \centering
  \caption{Number of bits per parameters of various types}
  \begin{tabular}{ll}
    \toprule
    TypeName & bit per parameter (bits) \\
    \midrule
    $\langle\QINT_{8}(2)\,\text{BF16/64/max}\rangle$ &  $8 + 16/64 = 8.25$ \\
    $\langle\text{MXFP8 (E4M3) int8/32/mx}\rangle$ &  $8 + 8/32 = 8.25$ \\
    $\langle\QINT_{6}(2)\,\text{BF16/64/max}\rangle$ &  $6 + 16/64 = 6.25$ \\
    $\langle\text{MXFP6 (E2M3) int8/32/mx}\rangle$ &  $6 + 8/32 = 6.25$ \\
    \bottomrule
  \end{tabular}
  \label{tab:bit_per_param}
\end{table}

In this way we can fairly compare the results between MX and AIQ
schemes. In table ~\ref{tab:THEORETIAL_COMPRESSION} we give the
expected compression levels for various real quantization scenario at
the 6.25 and 8.25 bit levels. In Tables~\ref{tab:AIQ_BENCHMARKS}
and~\ref{tab:MX_BENCHMARKS}, we report data collected in our fake
quantization experiments. Each cell reports the baseline (unquantized)
accuracy alongside the post-quantization accuracy for a given
model-benchmark pair. The bottom rows of each sub-table report the
average accuracy across all four benchmarks for both baseline and
quantized configurations, followed by the percentage difference
between the two. A positive difference indicates that the quantized
model matched or exceeded baseline performance, while a
negative difference indicates degradation.

\begin{table}
\centering
\caption{Theoretical Compression of All Linear Layers}
\label{tab:THEORETIAL_COMPRESSION}
\resizebox{\columnwidth}{!}{%
\footnotesize
\begin{tabular}{@{}lrrr@{}}
\toprule
Model & Baseline 16 Bit & 8.25 Bit & 6.25 Bit \\
\midrule
Llama-3.1-70B & 131.4 GB & 68.7 GB & 52.5 GB\\
Llama-3.1-8B & 15.0 GB & 8.2 GB  & 6.4 GB\\
Llama-3.2-1B & 2.3 GB & 1.2 GB  & 0.9 GB\\
Qwen2.5-7B & 14.2 GB & 7.8 GB & 6.2 GB\\
Mixtral-8x7B & 87.0 GB & 45.0 GB & 34.1 GB\\
\bottomrule
\end{tabular}
}
\end{table}

\begin{table}
\centering
\caption{AIQ Benchmark Accuracy (Baseline / Quantized)}
\label{tab:AIQ_BENCHMARKS}
\resizebox{\columnwidth}{!}{%
\footnotesize
\setlength{\tabcolsep}{3pt}
\begin{tabular}{lccccc}
\multicolumn{6}{@{}c@{}}{$\langle\QINT_{8}(2)\,\text{BF16/64/max}\rangle$ Weights Only} \\[6pt]
Benchmark & Llama-3.1-70B & Llama-3.1-8B & Llama-3.2-1B & Qwen2.5-7B & Mixtral-8x7B \\
\midrule
winogrande & 79.24 / 78.85 & 74.03 / 73.72 & 60.22 / 60.54 & 71.11 / 71.03 & 76.16 / 76.01 \\
piqa & 83.13 / 83.30 & 80.14 / 79.76 & 74.37 / 74.70 & 79.65 / 79.76 & 82.59 / 82.43 \\
boolq & 87.74 / 87.92 & 83.85 / 83.88 & 64.01 / 64.07 & 86.39 / 86.33 & 85.41 / 85.50 \\
lambada\_openai & 75.55 / 75.70 & 73.20 / 73.18 & 61.96 / 61.85 & 69.24 / 69.59 & 78.11 / 78.30 \\
\midrule
\textbf{Averages} & 81.42 / 81.44 & 77.81 / 77.64 & 65.14 / 65.29 & 76.60 / 76.68 & 80.57 / 80.56 \\
\textbf{Difference} & \color{green} 0.03\% & \color{red} -0.17\% & \color{green} 0.15\% & \color{green} 0.08\% & \color{red} -0.01\% \\
\midrule
\midrule
\\
\multicolumn{6}{@{}c@{}}{$\langle\QINT_{6}(2)\,\text{BF16/64/max}\rangle$ Weights Only} \\[6pt]
Benchmark & Llama-3.1-70B & Llama-3.1-8B & Llama-3.2-1B & Qwen2.5-7B & Mixtral-8x7B \\
\midrule
winogrande & 79.24 / 79.48 & 74.03 / 73.95 & 60.22 / 59.51 & 71.11 / 70.96 & 76.16 / 76.16 \\
piqa & 83.13 / 83.41 & 80.14 / 79.87 & 74.37 / 73.99 & 79.65 / 79.16 & 82.59 / 82.59 \\
boolq & 87.74 / 87.89 & 83.85 / 83.91 & 64.01 / 62.66 & 86.39 / 86.30 & 85.41 / 85.38 \\
lambada\_openai & 75.55 / 75.90 & 73.20 / 72.77 & 61.96 / 61.46 & 69.24 / 68.91 & 78.11 / 77.70 \\
\midrule
\textbf{Averages} & 81.42 / 81.67 & 77.81 / 77.62 & 65.14 / 64.41 & 76.60 / 76.33 & 80.57 / 80.46 \\
\textbf{Difference} & \color{green} 0.26\% & \color{red} -0.18\% & \color{red} -0.73\% & \color{red} -0.27\% & \color{red} -0.11\% \\
\bottomrule
\end{tabular}
}
\end{table}

\begin{table}
\centering
\caption{MX Benchmark Accuracy (Baseline / Quantized)}
\label{tab:MX_BENCHMARKS}
\resizebox{\columnwidth}{!}{%
\footnotesize
\setlength{\tabcolsep}{3pt}
\begin{tabular}{lccccc}
\multicolumn{6}{@{}c@{}}{$\langle\text{MXFP8(E4M3) int8/32/mx}\rangle$ Weights Only} \\[6pt]
Benchmark & Llama-3.1-70B & Llama-3.1-8B & Llama-3.2-1B & Qwen2.5-7B & Mixtral-8x7B \\
\midrule
winogrande & 79.24 / 79.64 & 74.03 / 73.95 & 60.22 / 60.69 & 71.11 / 69.93 & 76.16 / 76.24 \\
piqa & 83.13 / 83.30 & 80.14 / 80.03 & 74.37 / 74.81 & 79.65 / 79.38 & 82.59 / 82.26 \\
boolq & 87.74 / 87.77 & 83.85 / 83.64 & 64.01 / 64.04 & 86.39 / 86.48 & 85.41 / 84.28 \\
lambada\_openai & 75.55 / 75.51 & 73.20 / 72.54 & 61.96 / 61.56 & 69.24 / 68.35 & 78.11 / 77.59 \\
\midrule
\textbf{Averages} & 81.42 / 81.56 & 77.81 / 77.54 & 65.14 / 65.28 & 76.60 / 76.03 & 80.57 / 80.09 \\
\textbf{Difference} & \color{green} 0.14\% & \color{red} -0.27\% & \color{green} 0.14\% & \color{red} -0.56\% & \color{red} -0.47\% \\
\midrule
\midrule
\\
\multicolumn{6}{@{}c@{}}{$\langle\text{MXFP6(E2M3) int8/32/mx}\rangle$ Weights Only} \\[6pt]
Benchmark & Llama-3.1-70B & Llama-3.1-8B & Llama-3.2-1B & Qwen2.5-7B & Mixtral-8x7B \\
\midrule
winogrande & 79.24 / 79.56 & 74.03 / 73.80 & 60.22 / 60.30 & 71.11 / 69.85 & 76.16 / 76.32 \\
piqa & 83.13 / 83.24 & 80.14 / 79.82 & 74.37 / 74.81 & 79.65 / 79.49 & 82.59 / 82.15 \\
boolq & 87.74 / 87.61 & 83.85 / 83.73 & 64.01 / 63.67 & 86.39 / 86.61 & 85.41 / 84.46 \\
lambada\_openai & 75.55 / 75.47 & 73.20 / 73.14 & 61.96 / 60.28 & 69.24 / 68.97 & 78.11 / 77.41 \\
\midrule
\textbf{Averages} & 81.42 / 81.47 & 77.81 / 77.62 & 65.14 / 64.77 & 76.60 / 76.23 & 80.57 / 80.08 \\
\textbf{Difference} & \color{green} 0.06\% & \color{red} -0.18\% & \color{red} -0.38\% & \color{red} -0.37\% & \color{red} -0.48\% \\
\bottomrule
\end{tabular}
}
\end{table}


\section{Quantization Experimental Discussion}
At 8.25 bits per parameter,
$\langle\QINT_{8}(2)\,\text{BF16/64/max}\rangle$ proves to be at least
as accurate as the industry-standard
$\langle\text{MXFP8(E4M3) int8/32/mx}\rangle$ format, and in several
cases measurably better. AIQ's average accuracy differences range from
$+0.15\%$ (Llama-3.2-1B) down to $-0.17\%$ (Mixtral-8x7B), with the
largest single-benchmark drop being only $0.38$ points (piqa on
Llama-3.1-8B). By contrast, MXFP8 exhibits more pronounced degradation
on three of the five models: Qwen2.5-7B loses $0.56\%$ in average
accuracy, Mixtral-8x7B loses $0.47\%$, and Llama-3.1-8B loses
$0.27\%$. The two formats perform comparably on the 70B and 1B
models, where both are essentially lossless. Overall, AIQ matches or
outperforms MXFP8 on four of five models at identical bit budgets,
demonstrating that algebraic integer quantization is fully competitive
with the MX format at 8 bits.

At 6.25 bits per parameter, both formats incur more noticeable
degradation, and the comparison is
closer. $\langle\QINT_{6}(2)\,\text{BF16/64/max}\rangle$ outperforms
$\langle\text{MXFP6(E2M3) int8/32/mx}\rangle$ on three of the five
models: Llama-3.1-70B ($+0.26\%$ vs $+0.06\%$), Qwen2.5-7B ($-0.27\%$
vs $-0.37\%$), and most notably Mixtral-8x7B ($-0.11\%$ vs
$-0.48\%$). The two formats are effectively tied on Llama-3.1-8B (both
at $-0.18\%$). The one case where MX holds a clear advantage is
Llama-3.2-1B, where AIQ drops by $0.73\%$ compared to MX's $-0.38\%$,
suggesting that this particular model's weight distribution may be
less well-suited to the $\QINT_{6}$ representative set. Nonetheless,
AIQ is competitive or superior in the majority of configurations at
this bit level.

\textbf{Quantization-induced accuracy gains.}  A notable observation
in the results is that several models exhibited small but consistent
accuracy \textit{improvements} after quantization. Although
counterintuitive, this phenomenon is well documented and is generally
attributed to a regularization effect. The inherent rounding in
quantization can act as a form of weight normalization. These results
should not be taken as a fact that AIQ can improve models but rather
that AIQ did not degrade these specific models on these specific
benchmarks.

\textbf{Broader implications.}  Taken together, these results
demonstrate that AIQ is a viable quantization format for LLM
inference. In a fair bit-for-bit comparison, AIQ achieves accuracy
comparable to the industry-standard MX format, with neither scheme
holding a consistent advantage across all models and benchmarks. Given
that the MX specification has already been accepted by the industry as
suitable for hardware implementation and production deployment, the
fact that AIQ achieves competitive accuracy at equivalent bit budgets
establishes a strong case for further development.

Several avenues remain for further optimization of AIQ. The
experiments presented here employed only the simplest possible
configuration. We used a basic max scaling method and the
straightforward algebraic basis $\sqrt{2}$. One of the fundamental
strengths of AIQ is that its algebraic basis can be chosen or tuned to
match the weight distribution of a specific model or even individual
layers within a model, an optimization degree of freedom that simply
does not exist in floating-point or MX formats. Exploring alternative
bases, block sizes, scale data types, and more sophisticated scaling
strategies could yield substantial improvements over the already
competitive results reported here. Furthermore, these experiments
relied entirely on fake quantization, meaning that all computation was
still performed in the model's native floating-point precision. In a
true deployment, AIQ would additionally benefit from its lossless
arithmetic properties. By contrast, MX formats accumulate
floating-point rounding errors at every arithmetic step. The co-design
of hardware and software to exploit this error-free computation
represents a promising direction that could unlock accuracy advantages
fundamentally beyond what fake quantization alone can demonstrate.

\section{Comparison Between Floating Point and Algebraic Integer Data Format and Directions for Future Research}
\label{sec:future_research}

In this section, we will outline the advantages and disadvantages of
the algebraic integers not as a quantization mechanism but as a data
format, aimed at representing real and (if needed) complex numbers. As
a data format, algebraic integers possess some unique features that
can be extremely beneficial in many applications.

\textbf{The representation of zero.} One of the most obvious problems
with floating point data formats is the inability to represent zero
directly, requiring a special encoding convention. In contrast, the
algebraic-integer data format naturally and exactly encodes zero as
the pair (0,0). This yields a data format that can approximate every
real number with a desired precision while representing zero exactly.

\textbf{Dynamic range considerations.} Consider the FP8 data
formats. The e4m3 variant offers a dynamic range of approximately
448 while the e5m2 variant jumps to approximately 57,344 with nothing in
between.
A single additional exponent bit causes an explosive increase
in dynamic range, which may be undesirable. In contrast, algebraic
integer quantization offers the freedom to choose the irrational
component of the data format. For example, one can use representations
of the form $a + b \cdot \varphi^k$, where $\varphi$ is the golden
ratio and $k$ can be tuned to accommodate the dynamic range needed for
a specific application, such as a range-bound layer in a neural
network.

\textbf{Subnormals.} This, perhaps, is the most subtle difference
between AINF and FP. In the case of FP, it is easy to calculate the
largest and the smallest non-zero (by absolute value) representable
real numbers. Calculating the largest representable number in AINF is
easy, but it is more subtle to compute the smallest (by absolute
value) non-zero representable number. For example, if we use an AINF of
the form $a+b\sqrt{2}$ with $(a,b)$ three-bit signed integers, the
smallest number happens to be $(-3,2)$ which corresponds to
$0.171573$. The problem of finding the smallest AINF representative is
closely related to the shortest vector problem in lattices. The key
insight is that when particularly small subnormals are needed for a
given format (say, a 6-bit AINF data format with $(3,3)$-bitlengths),
one can search for a quadratic integer $z$ such that the shortest
vector in the corresponding lattice is unusually short. For example,
in analyzing 6-bit AINF data formats, we discovered that choosing $z =
\sqrt{75} - 8$ (the positive root of $x^2 + 16x - 11 = 0$)


yields an unusually small non-zero representable value: the pair $(2, -3)$,
corresponding to a magnitude of only $0.0195$. This effectively adds
nearly 4 extra bits of dynamic range compared to algebraic integers of
the form $a + b\sqrt{2}$.

This is particularly important for low-bit AIQ, where a large number
of small weights are simply rounded to zero for lack of a better
representable value. Pre-scaling can mitigate this by preventing the
wholesale discarding of small weights, but it is not without
drawbacks. It wastes dynamic range and introduces substantial
approximation errors for large outliers. With AIQ, this problem is
alleviated by the ability to have very small non-zero representable
values, afforded by the freedom to choose the root of the quadratic
(or higher-order) polynomial that defines the data format.


\section{Conclusion}
\label{sec:conclusion}

In this article, we have reintroduced algebraic integer number formats
as a powerful framework for compressed computation and proposed
algebraic integer quantization (AIQ) as a practical quantization
scheme for large language model inference. We developed a matrix
multiplication engine that exploits the algebraic integer encoding to
perform multiply-accumulate operations in an error-free manner using
purely integer arithmetic, demonstrating that AINF is not only a
viable representational format but also one that can be computed on
efficiently with existing hardware. Our fake quantization experiments
demonstrated that AIQ preserves model accuracy at both 8.25 and 6.25
bits per parameter, retaining near-original performance. Furthermore,
at both bit levels AIQ achieved accuracy competitive with the widely
adopted MX format, with neither scheme holding a consistent advantage
across all model and benchmark pairs, though individual model
sensitivity varied. Given that the industry has broadly accepted the
accuracy trade-offs of MX formats as suitable for production
deployment, the fact that AIQ achieves competitive accuracy at
equivalent bit budgets is a strong indication of its viability and
warrants further investigation into its unique optimization degrees of
freedom.

Beyond these empirical results, we have outlined several structural advantages of the algebraic integer data format over conventional floating-point representations, including exact associativity and distributivity, signless encoding, exact representation of zero, and fine-grained control over dynamic range and subnormal density through the choice of algebraic basis. These properties are particularly consequential at low bitwidths, where every bit of encoding capacity matters.

The results presented here employ quadratic algebraic integers exclusively, and numerous avenues for improvement remain. These include the exploration of alternative quadratic bases tailored to specific model architectures, the optimization of block sizes and scale factor strategies, and the extension to multidimensional algebraic integer quantization, which our preliminary analysis suggests can achieve dramatically finer subnormal resolution. The combination of algebraic integers with residue number systems for hardware acceleration represents another promising but as yet unexplored direction. The main limitation of the algebraic integer data format is the relative cost of division and comparison operations. This leaves the format particularly suited to LLM inference, namely matrix multiplication, where only additions and multiplications are required.

We believe this work demonstrates that algebraic integer quantization offers a compelling and fundamentally distinct approach to model compression, one whose configurability and arithmetic properties open optimization opportunities that are simply unavailable in floating-point or fixed-point formats. The work presented here is entirely software-based. On the other hand, all of the previous uses of algebraic integer arithmetic in DSP and image compression have been aimed at special purpose VLSI designs. This is due to the fact that on hardware level, there are additional figures of merit, like power consumption and area complexity. The AIQ VLSI designs for these applications have demonstrated superb performance in terms of ultra low power consumption. After clarifying the specifics of algebraic integer quantization and implementation of matmul in this article, we plan to explore the hardware benefits of AIQ in a future article.

\section*{Acknowledgment}
The authors would like to thank Pascal Lambert and Laurent Giorgi (Lemurian Labs Research Group, France) for their valuable contributions and helpful discussions during the development of this work.

\bibliographystyle{IEEEtran}
\bibliography{AIQ}

@ARTICLE{1055583,
  author={Reed, I. and Treiu-Kien Truong},
  journal={IEEE Transactions on Information Theory}, 
  title={Convolutions over residue classes of quadratic integers}, 
  year=1976,
  volume=22,
  number=4,
  pages={468-475},
  doi={10.1109/TIT.1976.1055583}}

@ARTICLE{1163370,
  author={Dubois, E. and Venetsanopoulos, A.},
  journal={IEEE Transactions on Acoustics, Speech, and Signal Processing}, 
  title={The generalized discrete Fourier transform in rings of algebraic integers}, 
  year={1980},
  volume={28},
  number={2},
  pages={169-175},
  doi={10.1109/TASSP.1980.1163370}}

@ARTICLE{1057081,
  author={Cozzens, J. and Finkelstein, L.},
  journal={IEEE Transactions on Information Theory}, 
  title={Computing the discrete Fourier transform using residue number systems in a ring of algebraic integers}, 
  year={1985},
  volume={31},
  number={5},
  pages={580-588},
  doi={10.1109/TIT.1985.1057081}}

@INPROCEEDINGS{101978,
  author={Games, R.A. and Moulin, D. and Rushanan, J.J.},
  booktitle={Proceedings of the 32nd Midwest Symposium on Circuits and Systems,}, 
  title={{VLSI} design of an algebraic-integer signal processor}, 
  year={1989},
  volume={},
  number={},
  pages={808-812 vol.2},
  doi={10.1109/MWSCAS.1989.101978}}

@INPROCEEDINGS{1207684,
  author={Wahid, K.A. and Dimitrov, V.S. and Jullien, G.A.},
  booktitle={Proceedings 2003 16th IEEE Symposium on Computer Arithmetic}, 
  title={Error-free arithmetic for discrete wavelet transforms using algebraic integers}, 
  year={2003},
  volume={},
  number={},
  pages={238-244},
  doi={10.1109/ARITH.2003.1207684}}

@INPROCEEDINGS{1467642,
  author={Wahid, K. and Dimitrov, V. and Jullien, G.},
  booktitle={17th IEEE Symposium on Computer Arithmetic (ARITH'05)}, 
  title={Error-free computation of $8\times8$ {2D} {DCT} and {IDCT} using two-dimensional algebraic integer quantization}, 
  year={2005},
  volume={},
  number={},
  pages={214-221},
  doi={10.1109/ARITH.2005.20}}

@INPROCEEDINGS{681703,
  author={Dimitrov, V.S. and Jullien, G.A. and Miller, W.C.},
  booktitle={Proceedings of the 1998 IEEE International Conference on Acoustics, Speech and Signal Processing, ICASSP '98 (Cat. No.98CH36181)}, 
  title={A new {DCT} algorithm based on encoding algebraic integers}, 
  year={1998},
  volume={3},
  number={},
  pages={1377-1380 vol.3},
  doi={10.1109/ICASSP.1998.681703}}

@article{doi:10.1049/el:20030406,
author = {V. Dimitrov  and G.A. Jullien },
title = {Multidimensional algebraic-integer encoding for high performance implementation of {DCT} and {IDCT}},
journal = {Electronics Letters},
volume = {39},
issue = {7},
pages = {602-603},
year = {2003},
doi = {10.1049/el:20030406},

URL = {https://digital-library.theiet.org/doi/abs/10.1049/el%3A20030406},
eprint = {https://digital-library.theiet.org/doi/pdf/10.1049/el%3A20030406}
}

@Misc{MITRE_AIQ,
  author = 	 {R. A. Games and S. D. O'Neil and J. J. Rushanan},
  title = 	 {Algebraic integer quantization and conversion},
  howpublished = {Tech. Rep. RADC-TR-88-167 (MTR-10337)},
  month = 	 {Jul},
  year = 	 1988,
  note = 	 {MITRE Corporation, Bedford, MA, USA}
}

@techreport{mx_spec,
  author = {Bita Darvish Rouhani and Nitin Garegrat and Tom Savell and Ankit More and Kyung-Nam Han and Ritchie Zhao and Mathew Hall and Jasmine Klar and Eric Chung and Yuan Yu and Michael Schulte and Ralph Wittig and Ian Bratt and Nigel Stephens and Jelena Milanovic and John Brothers and Pradeep Dubey and Marius Cornea and Alexander Heinecke and Andres Rodriguez and Martin Langhammer and Summer Deng and Maxim Naumov and Paulius Micikevicius and Michael Siu and Colin Verrilli},
  title = {{OCP} Microscaling Formats ({MX}) Specification},
  institution = {Open Compute Project},
  year = 2023,
  month = sep,
  type = {Technical Report},
  version = {1.0},
  url = {https://www.opencompute.org/documents/ocp-microscaling-formats-mx-v1-0-spec-final-pdf}
}

@misc{zhou2025precisionneuralnetworkquantization,
      title={Precision Neural Network Quantization via Learnable Adaptive Modules}, 
      author={Wenqiang Zhou and Zhendong Yu and Xinyu Liu and Jiaming Yang and Rong Xiao and Tao Wang and Chenwei Tang and Jiancheng Lv},
      year={2025},
      eprint={2504.17263},
      archivePrefix={arXiv},
      primaryClass={cs.CV},
      url={https://arxiv.org/abs/2504.17263}, 
}

@misc{askarihemmat2022qregregularizationeffectsquantization,
      title={{QReg}: On Regularization Effects of Quantization}, 
      author={MohammadHossein AskariHemmat and Reyhane Askari Hemmat and Alex Hoffman and Ivan Lazarevich and Ehsan Saboori and Olivier Mastropietro and Sudhakar Sah and Yvon Savaria and Jean-Pierre David},
      year={2022},
      eprint={2206.12372},
      archivePrefix={arXiv},
      primaryClass={cs.CV},
      url={https://arxiv.org/abs/2206.12372}, 
}

@article{DBLP:journals/cacm/SakaguchiBBC21,
  author       = {Keisuke Sakaguchi and
                  Ronan Le Bras and
                  Chandra Bhagavatula and
                  Yejin Choi},
  title        = {WinoGrande: an adversarial winograd schema challenge at scale},
  journal      = {Commun. {ACM}},
  volume       = {64},
  number       = {9},
  pages        = {99--106},
  year         = {2021},
  url          = {https://doi.org/10.1145/3474381},
  doi          = {10.1145/3474381},
  bibsource    = {dblp computer science bibliography, https://dblp.org}
}

@inproceedings{DBLP:conf/aaai/BiskZLGC20,
  author       = {Yonatan Bisk and
                  Rowan Zellers and
                  Ronan Le Bras and
                  Jianfeng Gao and
                  Yejin Choi},
  title        = {{PIQA:} Reasoning about Physical Commonsense in Natural Language},
  booktitle    = {The Thirty-Fourth {AAAI} Conference on Artificial Intelligence, {AAAI}
                  2020, The Thirty-Second Innovative Applications of Artificial Intelligence
                  Conference, {IAAI} 2020, The Tenth {AAAI} Symposium on Educational
                  Advances in Artificial Intelligence, {EAAI} 2020, New York, NY, USA,
                  February 7-12, 2020},
  pages        = {7432--7439},
  publisher    = {{AAAI} Press},
  year         = {2020},
  url          = {https://doi.org/10.1609/aaai.v34i05.6239},
  doi          = {10.1609/AAAI.V34I05.6239},
  bibsource    = {dblp computer science bibliography, https://dblp.org}
}

@inproceedings{DBLP:conf/naacl/ClarkLCK0T19,
  author       = {Christopher Clark and
                  Kenton Lee and
                  Ming{-}Wei Chang and
                  Tom Kwiatkowski and
                  Michael Collins and
                  Kristina Toutanova},
  editor       = {Jill Burstein and
                  Christy Doran and
                  Thamar Solorio},
  title        = {{BoolQ}: Exploring the Surprising Difficulty of Natural Yes/No Questions},
  booktitle    = {Proceedings of the 2019 Conference of the North American Chapter of
                  the Association for Computational Linguistics: Human Language Technologies,
                  {NAACL-HLT} 2019, Minneapolis, MN, USA, June 2-7, 2019, Volume 1 (Long
                  and Short Papers)},
  pages        = {2924--2936},
  publisher    = {Association for Computational Linguistics},
  year         = {2019},
  url          = {https://doi.org/10.18653/v1/n19-1300},
  doi          = {10.18653/V1/N19-1300},
  bibsource    = {dblp computer science bibliography, https://dblp.org}
}

@inproceedings{DBLP:conf/acl/PapernoKLPBPBBF16,
  author       = {Denis Paperno and
                  Germ{\'{a}}n Kruszewski and
                  Angeliki Lazaridou and
                  Quan Ngoc Pham and
                  Raffaella Bernardi and
                  Sandro Pezzelle and
                  Marco Baroni and
                  Gemma Boleda and
                  Raquel Fern{\'{a}}ndez},
  title        = {The {LAMBADA} dataset: Word prediction requiring a broad discourse
                  context},
  booktitle    = {Proceedings of the 54th Annual Meeting of the Association for Computational
                  Linguistics, {ACL} 2016, August 7-12, 2016, Berlin, Germany, Volume
                  1: Long Papers},
  publisher    = {The Association for Computer Linguistics},
  year         = {2016},
  url          = {https://doi.org/10.18653/v1/p16-1144},
  doi          = {10.18653/V1/P16-1144},
  bibsource    = {dblp computer science bibliography, https://dblp.org}
}

@misc{dettmers2023qloraefficientfinetuningquantized,
      title={{QLoRA}: Efficient Finetuning of Quantized {LLMs}}, 
      author={Tim Dettmers and Artidoro Pagnoni and Ari Holtzman and Luke Zettlemoyer},
      year={2023},
      eprint={2305.14314},
      archivePrefix={arXiv},
      primaryClass={cs.LG},
      url={https://arxiv.org/abs/2305.14314}, 
}

@misc{uchino2025emulationcomplexmatrixmultiplication,
      title={Emulation of Complex Matrix Multiplication based on the Chinese Remainder Theorem}, 
      author={Yuki Uchino and Qianxiang Ma and Toshiyuki Imamura and Katsuhisa Ozaki and Patrick Lars Gutsche},
      year=2025,
      eprint={2512.08321},
      archivePrefix={arXiv},
      primaryClass={cs.DC},
      url={https://arxiv.org/abs/2512.08321}, 
}

@Article{KarOfm63,
  author =	 {A. Karatsuba and Y. Ofman},
  title =	 {Multiplication of Multidigit Numbers on Automata},
  journal =	 {Soviet Physics---Doklady},
  volume =	 7,
  number =	 7,
  month =	 jan,
  pages =	 {595--596},
  year =	 1963,
}

\end{document}